\documentclass{article}
\usepackage{spconf,amsmath,graphicx}
\graphicspath{ {images/} }

\title{Mitigating the Impact of Speech Recognition Errors on Chatbot using Sequence-to-sequence Model}
\name{Pin-Jung Chen$^*$, I-Hung Hsu$^*$, Yi-Yao Huang$^*$, Hung-Yi Lee \thanks{*These authors contributed equally.}}
\address{ Department of Electrical Engineering, National Taiwan University \\
\small{\{b02504086, b02901053, b02901042, hungyilee\}@ntu.edu.tw}}
\begin{document}
%
\maketitle
\begin{abstract}
\ \ \ \ \ We apply sequence-to-sequence model to mitigate the impact of speech recognition errors on open domain end-to-end dialog generation. We cast the task as a domain adaptation problem where ASR transcriptions and original texts are in two different domains. In this paper, our proposed model includes two individual encoders for each domain data and make their hidden states similar to ensure the decoder predict the same dialog text. The method demonstrates that the sequence-to-sequence model can learn the ASR transcriptions and original text pair having the same meaning and eliminate the speech recognition errors. Experimental results on Cornell movie dialog dataset demonstrate that the domain adaption system help the  spoken dialog system generate more similar responses with the original text answers.
\end{abstract}
\section{Introduction}

\ \ \ \ \ Chatbot, also called Conversational Agent or Dialog System, is the new big thing in social media and is quickly changing the way we interact with services. Currently there are two types of approaches to deal with this task. In \textit{retrieval-based models}, one may use some kind of heuristic to select an appropriate response from a predefined repository given the input context. The heuristic can simply be a rule-based algorithm, or can be as complex as deep neural networks\cite{lowe2015ubuntu}. By contrast, in \textit{generative models}, the response is generated word by word from scratch. The LSTM sequence-to-sequence model is one type of generative models, which maximizes the probability of generating a response given the previous dialog turn. This approach can be trained end-to-end and achieves state-of-the-art results on neural response generation tasks.

Despite the success of sequence-to-sequence models in dialog generation, there is no sequence-to-sequence model focusing on dealing with ASR errors in end-to-end spoken dialog systems. 
The main problem is that a spoken dialog system requires an automatic speech recognition (ASR) system to perform speech to text conversion, but the transcription inevitably includes ASR errors. 
For spoken dialog systems, \cite{Hakkani-tür06beyondasr} \cite{Tur02improvingspoken} used word confusion networks to estimate word confidence scores on ASR texts. However, there is no related work on sequence-to-sequence model to solve this problem. There are some other works in different research fields trying to address the ASR error issues. \cite{DBLP:journals/corr/YuLL16f} used deep neural networks to model the error probability on spoken text summarization and \cite{kumar2015error} applied the conditional random field (CRF) model on ASR error detection. These methods can only detect ASR errors and try to recover from them, but they do not make use of the semantic information.

In this paper, we investigate the effects of ASR errors for sequence-to-sequence models on spoken dialog systems. More specifically, we cast the task as a domain adaptation problem where ASR transcriptions and original texts are from two different domains. The goal of this task is to ensure that the dialog system can generate the same response given inputs from the two domains. Thus, we propose a dual-encoder sequence-to-sequence model, which manages to force the state vectors (i.e. the last hidden state of the encoder) of the original texts and ASR transcriptions resemble each other so that the decoder can generate similar responses. In addition, we create a new ASR transcription dataset based on Cornell Movie-Dialogs Corpus. Empirically, experiment results on this dataset illustrate the outstanding performance by using our dual-encoder sequence-to-sequence model. For the real-world scenario lacking ASR conversion data, we conduct experiments using different percentages of ASR transcription data to demonstrate that our model can still achieve decent results. We show that the domain adaptation approach can help the dialog system to apprehend the same semantic meaning between ASR transcriptions and original texts.

Our contributions are three-fold:
\begin{itemize}
  \setlength\itemsep{-0.3em}
  \item We are the first to formulate the ASR error issue on spoken dialog systems as a domain adaptation problem; 
  \item We show that our dual-encoder sequence-to-sequence model outperforms the original sequence-to-sequence model by a large margin;
  \item Our dual-encoder sequence-to-sequence model can learn the same semantic meaning of the inputs from two different domains and can be easily applied to other ASR related research topics.
\end{itemize}

\section{Related Work}
\hspace{0.398cm} Speech media analytics and applications have been achieved successfully with the basic approach of cascading ASR modules with text processing systems. This framework works well when the ASR accuracy is relatively high, but less sound when more challenging real-world scenarios are considered. Thus, ASR error management to ameliorate the end-to-end performance in such integrated system gets more and more attentions. In this section, we will introduce several methods that manage to deal with this problem in different applications. Then, we will discuss some related works on dialog systems and domain adaptation.

\subsection{Approaches in related applications}
 \ \ \ \ \ Different approaches are explored to address such problem in distinct applications.\cite{6289102} harnessed speech translation task by jointly learning ASR and machine translation to optimize bilingual evaluation understudy scores \cite{Papineni:2002:BMA:1073083.1073135} directly. This method alleviates the issue that the best ASR parameters on minimizing the traditional word error rate will only lead to sub-optimal performance. The same idea works for spoken content retrieval tasks. By modifying ASR training systems \cite{7114229}, the performance of retrieval-based models will be improved. However, all of these methods required modifying ASR modules. In this paper, we focus on mitigating the ASR errors based on a given ASR system. Without modifying ASR modules, interactive error recovery is applied to deal with errors in speech-to-speech translation system. \cite{kumar2015error} employed conditional random field (CRF) models to detect ASR errors and attempted to resolve them by eliciting user feedback. In abstractive headline generation task for spoken content, \cite{DBLP:journals/corr/YuLL16f} proposed a method about considering ASR errors as a probability distribution. The work applied an attentive RNN to incorporate ASR error parameters into the attention mechanism. 

\subsection{Methods for dialog systems}
\hspace{0.44cm} For dialog systems, numerous works managing to address ASR failures relied on spoken language understanding modules. Word Confusion Networks(WCNs) provided a tighter integration of ASR and language understanding, which took word confidence scores into consideration rather than simply using ASR one-best hypotheses \cite{Hakkani-tür06beyondasr} \cite{Tur02improvingspoken}. The application of WCNs improved the ASR error tolerance in spoken language understanding (SLU). 
\cite{6494264} demonstrated that jointly training for predicting the optimal word as well as the slot sequence would achieve significant improvement in both recognition and semantic tagging accuracy simultaneously. Furthermore, neural approaches based on word embedding techniques were utilized to measure ASR confidence and used as additional SLU features to augment the system performance \cite{DBLP:journals/corr/SimonnetGCEM17}. While abounding works focusing on spoken language understanding has hastened ASR failure management in modular dialog systems, ASR error handling in end-to-end chatbots is rarely seen. Hence, in this paper, we are trying to solve this problem under such scenario.

\subsection{Domain Adaptation}
\ \ \ \ \ There has been extensive prior works on domain transfer learning. Among them, most works focused on transferring deep neural network representations from a labeled source dataset to a target domain dataset. For example, \cite{DBLP:journals/corr/TzengHSD17} proposed an adversarial domain adaptation method which tried to minimize the distance between the source and the target domain feature mappings. The main concept of these works is to guide feature learning by minimizing the difference between the source and target feature distributions \cite{pmlr-v37-ganin15}\cite{DBLP:journals/corr/TzengHZSD14}\cite{DBLP:journals/corr/Long015}. Common methods were using Maximum Mean Discrepancy (MMD) \cite{5376} as loss to accomplish this purpose. MMD computes the norm of the difference between two domain means.
Choosing an adversarial loss to minimize the domain shift is another common approach. One example is to add a domain classifier which predicts the binary domain label of the inputs, then follow by a domain confusion loss that encourages the classifier to predict as closely as possible to a uniform distribution over binary labels \cite{DBLP:journals/corr/TzengHDS15}. Both methods provide measures to estimate the difference between two feature domain distributions and can be referenced as future works.

\section{Domain Adaptation on ASR errors}
\ \ \ \ \ We consider the ASR error issue as a domain adaptation problem. For the spoken dialog systems, users can choose to use text or speech as the dialog input. If users choose speech as the input, there will contain some \textit{background voice noise} and \textit{speech recognition uncertainty} on the input data. Accordingly, we can consider that the texts and ASR transcriptions have the same latent feature but are observed from different interfaces. Ideally, users will get the same response no matter whether the input is given by text or speech. For example, if the text is ``Hey, nice to see you again." and its ASR conversion is ``hey thanks to see you again", the dialog system has to generate the same answer, for instance: ``What's up?". In this scenario, we can use domain adaptation methods to solve the problem.

In unsupervised adaptation, source texts $X_{s}$ and responses $Y_{s}$ are drawn from a source distribution $p_s(x, y)$, while target transcriptions $X_{t}$ are drawn from a target distribution $p_t(x, y)$, where there are no response observations.

Our goal is to learn a target representation, $M_{t}$ and a decoder $D_{t}$ that can correctly predict the answer at test time. Because of the ASR errors, direct supervised learning on the target domain do not work well. Domain adaptation methods instead learn a source representation mapping, $M_{s}$, along with a source decoder $D_{s}$, and then adapt that model to the target domain.

The main objective is to regularize the learning of the source and target mappings, $M_{s}$ and $M_{t}$, in order to minimize the distance between the source and target mapping distributions: $M_{s}(X_{s})$ and $M_{t}(X_{t})$. If the distance is close enough, then the source decoder, $D_{s}$, can be directly applied to the target representations, $M_{t}(X_{t})$, eliminating the need to learn a separate target decoder $D_{t}$.

To implement this concept, we propose the dual-encoder sequence-to-sequence model. In Section 4, we describe the ideas of our model and the approach used to minimize the distance between source and target mapping distributions. In Section 5, we introduce the ASR transcription dataset synthesized by ourselves based on Cornell Movie-Dialogs Corpus. In Section 6, we conduct three different experiments to prove that our problem formulation is appropriate, and to compare a common domain adaptation method with our dual-encoder sequence-to-sequence model.

\begin{figure*}[t!]
\centering
\includegraphics[width=1\textwidth]{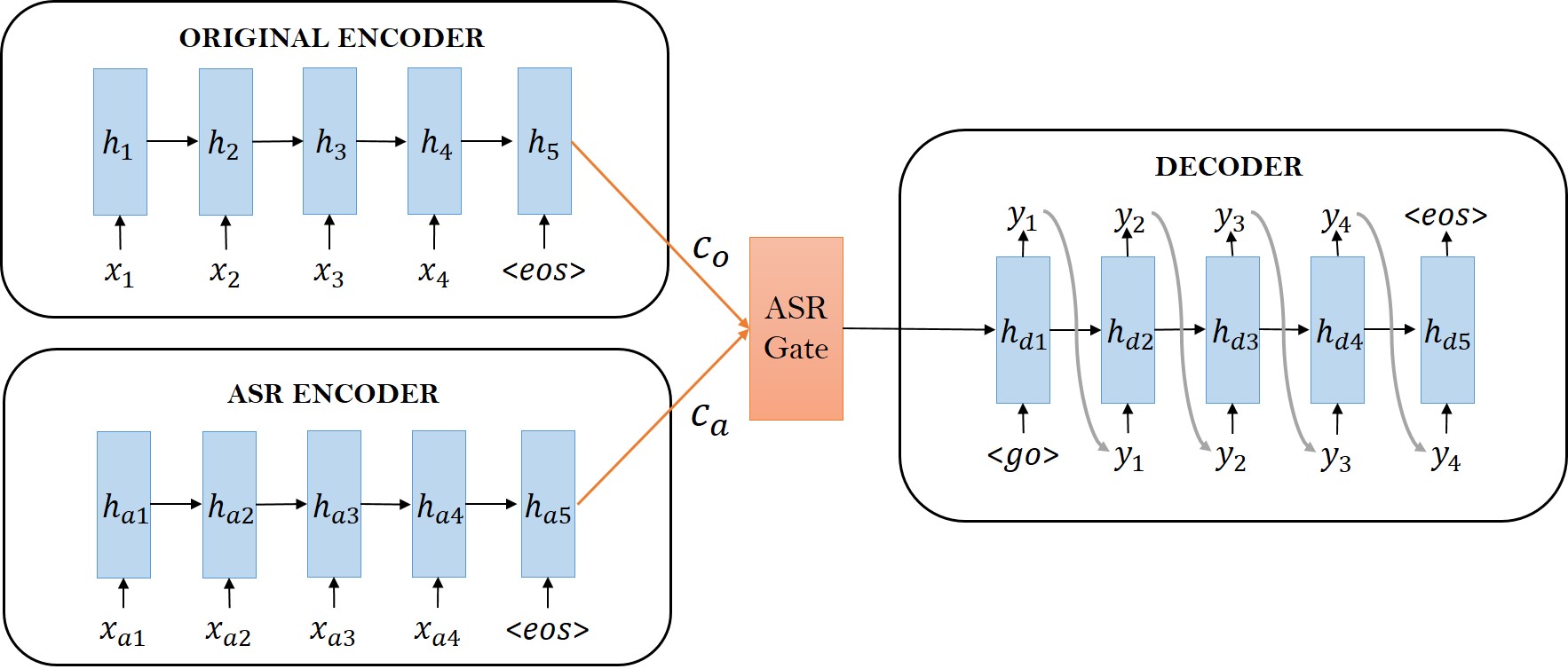}
\caption{The architecture of our dual-encoder sequence-to-sequence 
model for training the ASR transcription dialog system.}
\label{fig:arch}
\vspace{-2ex}
\end{figure*}

\section{Dual-Encoder Sequence-to-Sequence Model}
\ \ \ \ \ Given an ASR transcription $t_a$ or an original text $t$, our model will return a response $t_o$ predicted by the decoder. The dialog generation task is to ensure that the model can output similar $t_o$ from the $t_a$ and $t$ pair. We start from sequence-to-sequence model which consists of two recurrent neural networks (RNNs): an encoder that encodes the input and a decoder that generates the output. For our proposed model, it includes one decoder, one ASR gate and two encoders: one for ASR transcriptions and the other for original texts. Our dual-encoder sequence-to-sequence model is diagrammed in Figure~\ref{fig:arch}. 
\subsection{Encoder}

\ \ \ \ \ There are two encoders with the same structure but different parameters: one for ASR transcriptions, the other for original texts. The encoder is a RNN that reads each symbol of an input sequence $x = (x_1, x_2, x_3, \ldots)$  sequentially. The input sequence can be an ASR transcription or an original text. As it reads each symbol, the hidden state $h_t$ is updated by previous hidden state $h_{t-1}$ and $x_t$ according to Eq. (\ref{eq:1}).
\begin{equation} \label{eq:1}
h_t = f(h_{t-1}, x_t)
\end{equation}
After reading to the end of the sequence, the hidden state of the RNN is a summary $c$ of the whole input sequence. Both the summary $c_a$ from the ASR encoder and the summary $c_o$ from the original encoder will be passed to the ASR gate.

\subsection{ASR Gate}
\ \ \ \ \ Since the ASR transcription and the original text are observed from different domains but have the same semantic meaning, we expect the summary $c_o$ and $c_a$ to be the same. According to this assumption, we utilize the objective function Eq. (\ref{eq:2}) to minimize the distance between the two vectors in the training stage. In the testing stage, an ASR transcription will be encoded into summary $c_a$ and the ASR gate will forward it to the decoder.
\begin{equation}\label{eq:2}
L_c = \|c_o - c_a\|^2
\end{equation}

\subsection{Decoder}
\ \ \ \ The decoder of the proposed model is another RNN. It is trained to generate the output sequence by predicting the next symbol $y_t$ given the hidden state $h_t$.  Both $y_t$ and $h_t$ are also conditioned on $y_{t-1}$ and the summary $c$ selected by the ASR gate. Hence, the hidden state of the decoder at time $t$ is formulated as the following equation:
\begin{equation}
h_t = f(h_{t-1}, y_{t-1}, c)
\end{equation}
There are two phases in the training stage. In the first phase, original texts are provided and the ASR gate will pass summary $c_o$ to the decoder. The original encoder and the decoder are jointly trained to minimize the loss function Eq. (4).
\begin{equation}
L_s = - \frac{1}{N}\sum_{n=1}^{N}\log p_\theta(y_n|x_n)
\end{equation}
In the second phase, both original texts and ASR transcriptions are provided. We will fix parameters of the original encoder and forward summary $c_a$ to the decoder. The overall loss function $L$ of our dual-encoder sequence-to-sequence model in phase 2 is:
\begin{equation}
L = L_c + L_s
\end{equation}

\section{Dataset}
\ \ \ \ \ We use Cornell Movie-Dialogs Corpus \cite{Danescu-Niculescu-Mizil+Lee:11a} as our training and testing dataset. This corpus contains a large meta-data collection of fictional conversations extracted from raw movie scripts. It has 220,579 exchanges between 10,292 pairs of movie characters, with total 304,713 utterances. As turn taking is clearly indicated in this corpus, we can easily collect pairs of sentences as the data for our chatbot.

We perform only some basic pre-processing on this corpus. First, we normalize all digits to `0' and lowercase every letter. Second, we remove the noisy tags such as \textless u\textgreater \ and \textless /u\textgreater \ which are meaningless. Finally, since we adopt the bucketing method, we only retain those data which the original input sentences and their corresponding ASR transcriptions can be fit into the same bucket. After these pre-processing steps, we obtain 64731 turns as training data and 17853 turns as testing data.

Since currently there is no available corpus for spoken dialog systems, we do the following procedure to generate our ASR data. First, we use Google text-to-speech system to transform the Cornell Movie-Dialogs Corpus into audio files. Then, we utilize CMU Sphinx to generate the ASR transcriptions. We compare the ASR transcriptions to source texts using BLEU scores to analyze the extent to which the ASR errors affect our corpus as Table \ref{table:1}.

\begin{table}
\begin{center}
 \begin{tabular}{l c c} 
 \textbf{BLEU Score range} & \textbf{\# of data} \\
 \hline
 0 $\sim$ 0.4 & 25776  \\
 \hline
 0.4 $\sim$ 0.7 & 21568 \\
 \hline
 0.7 $\sim$ 1.0 & 17387 \\
 \hline
\end{tabular}
\caption{The ASR error distribution on our training dataset (higher BLEU scores indicate less ASR errors).}
\label{table:1}
\vspace{-2ex}
\end{center}
\end{table}

To be more specific, our dataset contains six parts:
\begin{itemize}
  \setlength\itemsep{-0.3em}
  \item Original texts for training data input (\textbf{train.enc})
  \item Original texts for training data output (\textbf{train.dec})
  \item ASR transcriptions for training data input (\textbf{train\_asr.enc})
  \item Original texts for testing data input (\textbf{test.enc})
  \item Original texts for testing data output (\textbf{test.dec})
  \item ASR transcriptions for testing data input (\textbf{test\_asr.enc})
\end{itemize}
There is no ASR transcription for training and testing data outputs because the training and testing data outputs of ASR transcriptions are the same as \textbf{train.dec} and \textbf{test.dec} respectively. For the expression to be clear, we will use the boldface notation (\textbf{train.enc}, \textbf{train.dec}, etc\ldots) in Section 6.

\section{Experiments}

\subsection{Implementation}
\hspace{0.51cm} Our algorithm is implemented in Tensorflow. We conduct three different experiments to show the ASR dialog problem and to prove that our model can address this issue. 
The original encoder, ASR encoder and decoder of our Seq2seq model are GRU cells with two layers of 512 dimensions. The buckets are [(10, 10), (20, 20)]. Adam optimizer is chosen \cite{journals/corr/KingmaB14} to optimize our network with details presented in Table \ref{table:parameter}. Gradients are clipped to avoid gradient explosion with a threshold of 5. The vocabulary size is limited to 35000, and the word embedding size is 64 with random initialization. During the training process, we feed the ground truth to the next time step of the decoder. The training details are summarized in Table \ref{table:parameter}.

Besides, we've tried the scheduled sampling \cite{DBLP:journals/corr/BengioVJS15} mechanism, but it does not have any positive effect on decoding more readable sentences even if it can reduce the gap between training and testing perplexity. Thus, we refrain from using scheduled sampling in our final model.

\begin{table}[h!]
\small
\centering
\begin{tabular}{ |c|c| } 
 \hline
 GRU layer dimensions & 512 \\
 Batch size \textit{B} & 64 \\
 Initial learning rate $\lambda$ & 0.002 \\
 $\beta_1$ & 0.9 \\
 $\beta_2$ & 0.999 \\
 epsilon & 1e-8 \\
 vocabulary size & 35000 \\
 word embedding size & 64 \\
 \hline
\end{tabular}
\caption{Parameter settings}
\vspace{-1ex}
\label{table:parameter}
\end{table}

\subsection{End-to-end training based on ASR transcriptions}
\hspace{0.398cm} This experiment aims to demonstrate that the model directly trained on ASR transcriptions will generate diverse dialogs compared to the model trained on original texts. There are two sequence-to-sequence models being trained: 
\begin{itemize}
  \setlength\itemsep{-0.3em}
  \item \textbf{Seq2Seq-text}: sequence-to-sequence model trained on original text.
  \item \textbf{Seq2Seq-ASR}: sequence-to-sequence model trained on ASR transcriptions.
\end{itemize}

We denote \textbf{test.pred} as the responses predicted by model \textbf{Seq2Seq-text} given \textbf{test.enc}. The two models, \textbf{Seq2Seq-text} and \textbf{Seq2Seq-ASR}, are then tested by inputting data with ASR errors (\textbf{test\_asr.enc}). We examine whether the two models can generate responses similar to \textbf{test.pred}, and the results are shown in Table \ref{table:3}.

\begin{table}[t]
\begin{center}
 \begin{tabular}{l c c} 
 \textbf{Model} & \textbf{Training data} & \textbf{BLEU Score} \\
 \hline
 \textbf{Seq2Seq-text} & original text & 0.1255  \\ 
 \textbf{Seq2Seq-ASR} & ASR transcription & 0.1812  \\ 
 \hline
\end{tabular}
\caption{Compared to \textbf{test.pred}, the result on \textbf{test\_asr.enc} using the end-to-end training method. }
\label{table:3}
\vspace{-1ex}
\end{center}
\end{table}

According to the results, end-to-end training on ASR transcriptions can get better performance than the \textbf{Seq2Seq-text} which is only trained on original texts. However, both of the results are lower than 0.2. It indicates that these approaches can not guarantee dialog systems to reply a similar response when giving an ASR transcription input. Our explanation is that the model trained on ASR transcriptions can make out some ASR patterns and thus outperforms \textbf{Seq2Seq-text}. Nevertheless, the ASR patterns are quite inconsistent, causing the model to be incapable of learning them well.

\subsection{Fine-tune on original text encoder}
\hspace{0.44cm} To solve the ASR error problem, we experiment with a common domain algorithm: fine-tuning \textbf{Seq2Seq-text} on ASR transcriptions. In this method, the model can learn the semantic meanings of original texts first and then adapt itself to data from the ASR domain. For the real world scenario, the original text data are much more than the ASR transcription data. Therefore, we fine-tune our model on different percentages of \textbf{train\_asr.enc} to examine its performance when fewer data are available. The result is in Table \ref{table:4}.

\begin{table}[t]
\begin{center}
 \begin{tabular}{l c c} 
 \textbf{Model} & \textbf{Fine-tuning data} & \textbf{BLEU Score} \\
 \hline
 \textbf{Seq2Seq-text} & 0\% train\_asr.enc  & 0.1255 \\ \cline{2-3}
                                & 20\% train\_asr.enc & 0.1763  \\ \cline{2-3}
                                & 40\% train\_asr.enc & 0.1916  \\ \cline{2-3}
                                & 60\% train\_asr.enc & 0.2004  \\ \cline{2-3}
                                & 80\% train\_asr.enc & 0.2037  \\ \cline{2-3}
                                & 100\% train\_asr.enc & 0.2016  \\ \hline
\end{tabular}
\caption{Compared to \textbf{test.pred}, the result on \textbf{test\_asr.enc} using different percentages of ASR data to fine-tune. }
\label{table:4}
\vspace{-1ex}
\end{center}
\end{table}

Our result shows that the performance becomes significantly better when training on more ASR data. Also, on the 40\% \textbf{train\_asr.enc} experiment, the performance already outperforms the \textbf{Seq2Seq-ASR} result. It can prove our assumption: ASR dialog systems can be viewed as a domain adaptation problem of text dialog systems. However, ASR errors include incorrectly recognized words as well as the removal of punctuation marks. Consequently, the structures of ASR transcriptions and original texts will be different in essence, even though they have the same meaning. Therefore, we propose the dual-encoder sequence-to-sequence model to solve this problem and it achieves the best result. 

\subsection{Dual-encoder sequence-to-sequence model}
\ \ \ \ \ In this experiment, we train \textbf{Seq2Seq-text} first and then build another encoder for ASR transcription inputs to minimize the state vectors of the two encoders by Eq. (2). For testing, \textbf{test\_asr.enc} is used as the encoder input. The ASR gate will allow the ASR state vector to be passed into the decoder. The result is shown in Table \ref{table:5}. 
\begin{table}[t]
\begin{center}
 \begin{tabular}{l l c c} 
 Loss functions & \textbf{Training data} & \textbf{BLEU Score} \\
 \hline
 $L_c$ & 20\% train\_asr.enc & 0.2560  \\  \cline{2-3}
               & 40\% train\_asr.enc & 0.2567  \\  \cline{2-3}
               & 60\% train\_asr.enc & 0.2639  \\  \cline{2-3}
               & 80\% train\_asr.enc & 0.2649  \\  \cline{2-3}
               & 100\% train\_asr.enc & \textbf{0.2692}  \\ \cline{2-3}
 \hline
 \hline
 $L_c+L_s$ & 20\% train\_asr.enc &  0.2554  \\  \cline{2-3}
               & 40\% train\_asr.enc & 0.2643 \\  \cline{2-3}
               & 60\% train\_asr.enc & 0.2663  \\  \cline{2-3}
               & 80\% train\_asr.enc & 0.2683  \\  \cline{2-3}
               & 100\% train\_asr.enc & \textbf{0.2752}  \\ \cline{2-3}
 \hline
\end{tabular}
\caption{Compared to \textbf{test.pred}, the results of our \textbf{dual-encoder sequence to sequence} model on \textbf{test\_asr.enc}. }
\label{table:5}
\vspace{-1ex}
\end{center}
\end{table}

\begin{table*}[t]
\begin{center}
 \begin{tabular}{l l l c c} 
 \textbf{Affecting factors} & \textbf{Input} & \textbf{Sequence-to-Sequence model} & \textbf{Our model} \\
 \hline
 Punctuation Marks & \textbf{\textit{How old are you?}} & Thirty-five. & Twenty-five.  \\ 
 & \textbf{\textit{How old are you}} & Very strange. & Thirty-five.  \\
 \cline{2-4}
 & \textbf{\textit{What's your name?}} & Palm Smith. & Samantha.  \\ 
 & \textbf{\textit{What is your name}} & You're a good idea. & Jacob.  \\
 \hline
  \hline
 ASR error & \textbf{\textit{Do you like me?}} & Sure. & Yes.  \\
 & \textbf{\textit{The u. like me}} & What sub? & Yes.  \\ 
 \cline{2-4}
 & \textbf{\textit{How many days are you gonna stay?}} & Two. & I don't know.  \\ 
 & \textbf{\textit{How many days are you then as they}} & Why? & Twelve.  \\
 \hline
 \hline
 Homonyms & \textbf{\textit{How about going home?}} & Okay. & Okay.  \\
 & \textbf{\textit{How a bell going home?}} & I like the noise. & Okay.  \\ 
 \cline{2-4}
 & \textbf{\textit{I'm walking along the river.}} & You're a fool! & You're not!  \\
 & \textbf{\textit{I'm walking a long the river.}} & I'm sorry, I'm sorry. & What's the matter? \\
 \hline
\end{tabular}
\caption{Different types of affecting factors we observed from our positive examples.}
\label{table:6}
\vspace{-2ex}
\end{center}
\end{table*}

According to the Table \ref{table:5}, we can observe that all the training settings in this experiment achieve better performance than the previous one. The model gets 0.25 BLEU score by only using 20\% \textbf{train\_asr.enc} and its original texts pair for training. It can be applied to the real world scenario lacking ASR transcription data. With more ASR data, the performance gradually improves and reaches the best BLEU score of 0.269 when utilizing all \textbf{train\_asr.enc}.

Our dual-encoder sequence-to-sequence model deals with this task in two aspects. First, if the original text encoder is not capable of learning the meanings and structures of ASR data, we can use two distinct encoders to train the two domain data independently. Second, the method forces the state vector pair to be as similar as possible, which ensure the decoder to predict the same answer. In this way, the dialog system can give consistent responses for the text input or the speech input.

To be more advanced, we combine the two loss functions to optimize our model. It means that we also fine-tune the decoder on the target domain, in addition to minimizing the state vectors of the encoders. The result is in Table \ref{table:5}.

\subsection{Discussion}
\hspace{0.44cm} To understand how our model recovers from the ASR errors and generates suitable responses, we analyze some examples and observe that our model mitigates the impact of the noise in three aspects. The samples are shown in Table \ref{table:6}.
\subsubsection{Punctuation Marks}
\hspace{0.398cm} For the ASR transcriptions, there are no punctuation marks in every sentence. However, punctuation marks extremely influence the performance of the sequence-to-sequence model on the original text domain. In English sentences, punctuation marks are used to express our emotion. With only few changes in punctuation marks, the response will be totally different. Thus, the first challenge for our dual-encoder sequence-to-sequence model is to reduce the effect of lacking punctuation marks. For example, if the input is ``How old are you" without question marks, the sequence-to-sequence model responses an irrelevant answer ``Very strange.". For our proposed method, it replies ``Thirty-five.", which is an appropriate answer.
\subsubsection{ASR errors}
\hspace{0.4cm} ASR errors are caused by background noise and recognition system uncertainty. Our model learns to capture the semantic meanings of dialogs so it is less susceptible to ASR errors. For instance, when the input sentence ``How many days are you gonna stay?" is wrongly transferred into ``How many days are you then as they", the sequence-to-sequence model generates a meaningless answer ``Why?". On the contrary,  dual-encoder sequence-to-sequence model can still grasp the meaning and then replies ``Twelve".
\subsubsection{Homonyms}
\hspace{0.5cm} The difficulty in dealing with homonyms is another main issue on ASR systems. For example, ``about" may be wrongly recognized as ``a bell" so that the sequence-to-sequence model replies ``I like the noise." because of the word ``bell". However, our model responses ``Okay" for both inputs.

In our experiment, we have two important observations. (1) We prove that the effect of ASR errors can be mitigated by using several domain adaptation approaches. (2) Our dual-encoder sequence-to-sequence model achieves significant improvements on spoken dialog systems and can tolerate several types of noise.

\section{Conclusion}
\ \ \ \ In this paper, we propose a dual-encoder sequence-to-sequence model with two distinct loss functions to mitigate the effect of ASR errors on chatbots. Experimental results indicate that our domain adaptation concept is capable of minimizing the distance between the two feature mappings in each domain. The sample analysis also demonstrates that our model is reasonably robust on several types of noise.

\newpage
\label{sec:ref}

\bibliographystyle{IEEEbib}
\bibliography{strings, refs}

\end{document}